\documentclass[sigconf]{acmart}

\usepackage{enumitem}
\usepackage{booktabs}
\usepackage{multirow, makecell}

\AtBeginDocument{%
  }

\copyrightyear{2025}
\acmYear{2025}
\setcopyright{cc}
\setcctype{by}
\acmConference[FMSys '25]{The 2nd International Workshop on Foundation Models for Cyber-Physical Systems \& Internet of Things }{May 6--9, 2025}{Irvine, CA, USA}
\acmBooktitle{The 2nd International Workshop on Foundation Models for Cyber-Physical Systems \& Internet of Things (FMSys '25), May 6--9, 2025, Irvine, CA, USA}
\acmDOI{10.1145/3722565.3727198}
\acmISBN{/2025/05}

\begin{document}

\title{Toward Foundation Models for Online Complex Event Detection in CPS-IoT: A Case Study}

\author{Liying Han}
\email{liying98@ucla.edu}
\affiliation{%
  \institution{UCLA}
  \city{}
  \state{}
  \country{}
}

\author{Gaofeng Dong}
\email{gfdong@g.ucla.edu}
\affiliation{%
  \institution{UCLA}
  \city{}
  \state{}
  \country{}
}

\author{Xiaomin Ouyang}
\authornote{This work was done while the author was at UCLA.}
\email{xmouyang@cse.ust.hk}
\affiliation{%
  \institution{HKUST}
  \city{}
  \country{}
}

\author{Lance Kaplan}
\email{lance.m.kaplan.civ@army.mil}
\affiliation{%
 \institution{DEVCOM Army Research Laboratory}
 \city{}
 \state{}
 \country{}}

\author{Federico Cerutti}
\email{federico.cerutti@unibs.it}
\affiliation{%
  \institution{University of Brescia}
  \city{}
  \state{}
  \country{}}

\author{Mani Srivastava}
\authornote{The author holds concurrent appointments as an Amazon Scholar and a Professor at UCLA, but the work in this paper is unrelated to Amazon.}
\email{mbs@ucla.edu}
\affiliation{%
  \institution{UCLA}
  \city{}
  \state{}
  \country{}
}

\renewcommand{\shortauthors}{Han et al.}

\begin{abstract}
\emph{Complex events} (\emph{CE}s) play a crucial role in CPS-IoT applications, enabling high-level decision-making in domains such as smart monitoring and autonomous systems. However, most existing models focus on short-span perception tasks, lacking the long-term reasoning required for \emph{CE} detection. \emph{CE}s consist of sequences of short-time \emph{atomic events} (\emph{AE}s) governed by spatiotemporal dependencies. Detecting them is difficult due to long, noisy sensor data and the challenge of filtering out irrelevant \emph{AE}s while capturing meaningful patterns. This work explores \emph{CE} detection as a case study for CPS-IoT foundation models capable of long-term reasoning. We evaluate three approaches: (1) leveraging large language models (LLMs), (2) employing various neural architectures that learn \emph{CE} rules from data, and (3) adopting a neurosymbolic approach that integrates neural models with symbolic engines embedding human knowledge. Our results show that the state-space model, Mamba, which belongs to the second category, outperforms all methods in accuracy and generalization to longer, unseen sensor traces. These findings suggest that state-space models could be a strong backbone for CPS-IoT foundation models for long-span reasoning tasks.

\end{abstract}




\maketitle

\begin{figure*}[t]
    \centering
\includegraphics[width=0.95\linewidth]{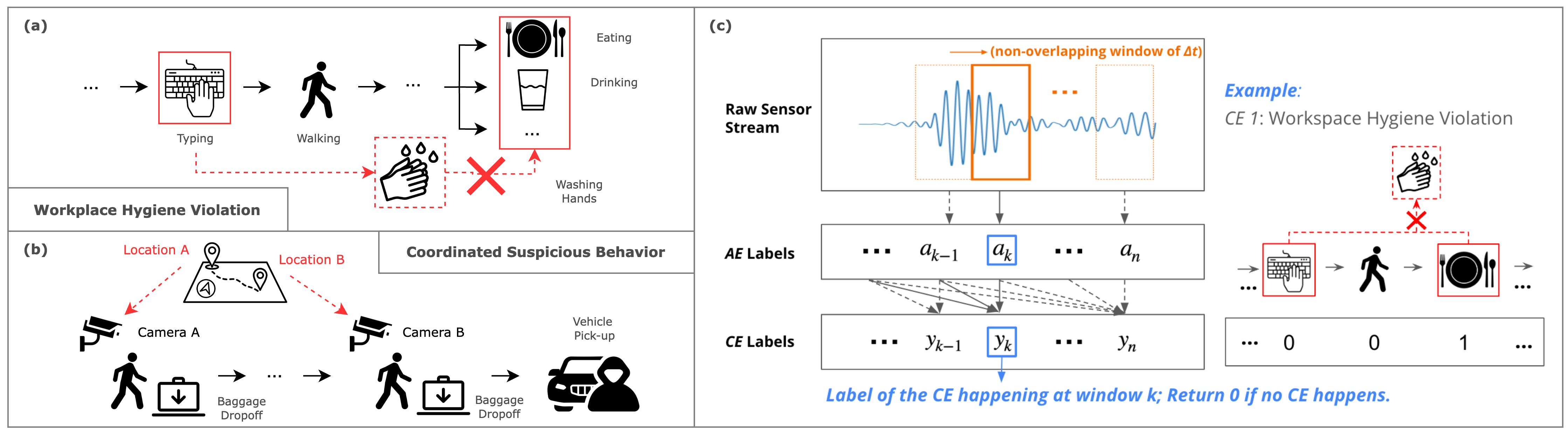}
\caption{(a) An intelligent smartwatch assistant monitors workplace hygiene and alerts users of potential violations. (b) In a smart facility, a surveillance system detects suspicious activities, such as unusual parcel hand-offs, using distributed cameras. (c) An illustration of the online CED task showing sensor processing and \emph{CE} labeling.
}
\Description{}
\label{fig:ce_overview}
\vspace{-1em}
\end{figure*}

\section{Introduction}
Current CPS-IoT applications excel in short-span perception tasks like human activity recognition and object detection, which require only a few seconds of sensor data. However, real-world scenarios often demand high-level contextual understanding over long periods for complex decision-making, a capability largely overlooked. For instance, an elderly person wearing a smartwatch for continuous health monitoring generates IMU data over hours. An 8-hour recording at 100 Hz produces 2.88 million data points, making it crucial to efficiently compress and retain key information to identify long-term, meaningful patterns related to health conditions.  

To effectively capture these long-term patterns, it is useful to represent them as \emph{\textbf{Complex Events}} (\emph{\textbf{CE}}s), which consist of sequences of short-time \emph{\textbf{Atomic Events}} (\emph{\textbf{AE}}s) governed by spatiotemporal dependencies. They are crucial for decision-making in CPS-IoT applications such as smart monitoring and autonomous systems that require long-term contextual understanding, as illustrated in Fig.~\ref{fig:ce_overview}(a) and (b). \textbf{Detecting \emph{CE}s is challenging} for three reasons. First, the model must identify relevant \emph{AE}s while ignoring irrelevant ones, increasing sequence-matching complexity. For instance, a sanitary protocol can be represented as ``Use restroom → X → Wash hands → X → Eat,'' where ``X'' includes irrelevant AEs like ``walking'' or ``sitting,'' complicating detection. Second, \emph{CE}s involve long-term dependencies with varying temporal gaps, making it difficult to model temporal and logical rules accurately. Third, many CPS-IoT applications require online detection for real-time responses to safety-critical events, necessitating efficient and accurate inference.  

Two main approaches exist for \emph{CE} detection. Data-driven methods use neural networks to learn patterns from data but struggle with long-term dependencies due to fixed context windows or inefficient memory mechanisms. For example, Recurrent Neural Networks (RNNs) and Transformers capture context through sequential processing and attention mechanisms but face memory fading and computational inefficiency over long durations. Recent advances in state-space models (SSMs), such as Mamba \cite{gu2022ssm,dao2024mamba2,gu2024mamba}, show promise by modeling event progression and capturing temporal dependencies. However, their effectiveness in noisy, high-dimensional sensor data for \emph{CE} detection remains unexplored.  

Neurosymbolic frameworks, as the second approach, combine neural networks with symbolic reasoning to leverage predefined rules for \emph{CE} patterns, reducing the need for extensive labeled data. However, symbolic engines are sensitive to noise, and probabilistic reasoning engines \cite{deepproblog,scallop,neurasp,slash}, although more robust through probabilistic programming, require complex temporal rule implementation and demand specialized expertise.  

Large Language Models (LLMs) are emerging as competitive alternatives due to their large-scale pre-training and enhanced reasoning abilities, enabling them to model long-term dependencies. Recent works like \cite{penai,llmsense} apply LLMs in sensor-based reasoning for CPS-IoT. However, their tasks do not address complex spatiotemporal dependencies or long temporal rule-based reasoning required for \emph{CE} detection.

This work explores online \emph{CE} detection as a case study for building CPS-IoT foundation models capable of long-term reasoning, enabling improved situational awareness for downstream tasks. We investigate three approaches: (1) leveraging LLMs as \emph{CE} detectors, (2) employing neural architectures that learn \emph{CE} rules directly from data, and (3) integrating neural models with symbolic engines embedding human knowledge. Our evaluation shows that the state-space model, Mamba, from the second category, outperforms all methods in both accuracy and generalization to longer, unseen sensor traces. These findings highlight the potential of state-space models as robust backbones for CPS-IoT foundation models designed for long-term complex event detection.  

\section{Related Works}
\textbf{Complex Event Detection (CED)} has been extensively studied in traditional stream processing for structured databases. Early works \cite{CEPoverview_2012,Schultz_2009,Debar_2001} employed event engines such as Finite State Machines (FSMs) to identify complex events based on predefined patterns. More recently, research has shifted towards CED over unstructured, high-dimensional data. Approaches such as \cite{xing2020neuroplex,ROIGVILAMALA2023119376} integrate neural detectors with symbolic rule-based systems, enabling backpropagation through logical rules. However, these methods suffer from computational scalability issues. Moreover, most existing approaches focus on complex activities within short time spans \cite{khan2023hybridgraphnetworkcomplex,activitynet,THUMOS14}, limiting their applicability to longer sensor traces with extended temporal dependencies.

\textbf{State-Space Models (SSMs)} capture long-span temporal dependencies through state transitions, maintaining state representations that implicitly encode temporal progression. Recent advances, such as Mamba \cite{gu2022ssm,dao2024mamba2,gu2024mamba}, show superior performance in long-term contextual understanding by using selection mechanisms to process sequences more efficiently and adaptively model different temporal dependencies. However, SSMs remain underexplored in noisy, high-dimensional sensor environments, particularly for long-span temporal events in CPS-IoT applications.
\section{Online Complex Event Detection}
\subsection{Complex Event Definitions}

\begin{definition}
\label{def:AE}
\emph{Atomic events} (\emph{AE}s) are short-duration, low-level events that serve as building blocks for complex events. They occur instantaneously or within a small time window and are directly detectable by models like image classification, object detection, or activity recognition.
\end{definition}

\begin{definition}
\label{def:CE}
\emph{Complex events} (\emph{CE}s) are high-level events defined as sequences or patterns of atomic events (\emph{AE}s) occurring in specific temporal or logical relationships.
\end{definition}
Let $A = {a_1, a_2, \ldots, a_n}$ be the set of all \emph{AE}s, where each $a_i$ has a start and end time. Similarly, let $E = {e_1, e_2, \ldots, e_k}$ be the set of all \emph{CE}s of interest.

Each complex event (\emph{CE}) $e_i \in E$ is defined as:
\[
e_i = R_i(A_i) = R_i(a_i^1, a_i^2, \ldots, a_i^{n_i}),
\]
where $A_i \subseteq A$ is the subset of \emph{AE}s relevant to $e_i$, $R_i$ is a \textit{pattern function} defining the temporal or logical relationship among the \emph{AE}s in $A_i$, and $n_i = |A_i|$ is the number of \emph{AE}s involved in defining $e_i$. Each $e_i$ has a time $t_{e_i}$, which is the moment (or interval) when the pattern $R_i$ is satisfied, indicating the occurrence of $e_i$.

\emph{Pattern Function ($R_i$)} maps $A_i$ to $e_i$ by defining patterns among the \emph{AE}s. We consider four main categories, some of which include subcategories:
\begin{enumerate}[leftmargin=1.5em, noitemsep]\label{def:pattern}
    \item \textbf{\emph{Sequential Patterns:}} \emph{Relaxed Order} – Key \emph{AE}s must occur in order but may include unrelated \emph{AE}s in between.
    \item \textbf{\emph{Temporal Patterns:}} (a) \emph{Duration Based} – Measuring the duration of specific \emph{AE}s; (b) \emph{Timing Relationship} – Capturing relative timing constraints, such as min and max delays between \emph{AE}s.  
    \item \textbf{\emph{Repetition Patterns:}} (a) \emph{Frequency Based} – Counting occurrences of \emph{AE}s over a time window; (b) \emph{Contextual Count} – Counting \emph{AE}s relative to other \emph{AE}s' timing.  
    \item \textbf{\emph{Combination Patterns:}} Merging \textbf{\emph{Sequential}} and \textbf{\emph{Temporal Patterns}} to express complex relationships.  
\end{enumerate}
Importantly, all patterns considered in this work are {\emph{bounded to finite states}, enabling representation by finite state machines (FSMs).

\subsection{Online Detection Task Formalization}\label{sec:CED-task}


Assume a system receives a raw data stream $\mathbf{X}$ from a single sensor with modalities $M$ at a sampling rate $r$. The system processes the stream using a non-overlapping sliding window of length $\Delta t$. At the $t$th window, the data segment is:
\begin{equation}
    \mathbf{D}_t=\mathbf{X}(t), \quad \mathbf{X}(t) \in \mathbb{R}^{(r \times \Delta t) \times m}
\end{equation}
where $m$ is the feature dimension of the sensor data of modality $M$.

Each window $t$ has a corresponding ground-truth \emph{CE} label $y_t$, which depends on the \emph{AE}s from previous windows $t-1$ and the current window $t$. As illustrated in Fig.~\ref{fig:ce_overview}(c), if a \emph{CE} spans from $t_1$ to $t_2$, then only $y_{t_2}$ is labeled as the \emph{CE}, while all $y_{t_1}$ to $y_{t_2 - 1}$ are labeled as ``0'' to indicate no \emph{CE} is detected before $t_2$.

For up to $T$ sliding windows, the objective of the real-time CED model $f$ is to minimize the difference between the predicted CE label $\hat{y_t}$ and the ground-truth label $y_t$ at every window:
\begin{equation}\label{eq:1}
    \min |\hat{y_t} - y_t|, \quad \textrm{ where }\hat{y_t} = f\left(\mathbf{D}_t\right), \quad 1 \leq t \leq T,
\end{equation}
This forms a \emph{multi-label multi-class classification} problem, where the objective can be expressed in vector form as:
\begin{equation}
    \min ||f(\mathbf{D}) - \mathbf{y}||, \quad \textrm{ where } \mathbf{D} = \left\{\mathbf{X}(1),\ldots, \mathbf{X}(T)\right\}.
\end{equation}
For data-driven methods, the task uses \textbf{\emph{high-level, coarse CE labels}} without fine-grained \emph{AE} labels during training, requiring the model to learn \emph{AE} semantics and \emph{CE} rules simultaneously. This represents a \textbf{unique} and \textbf{challenging} combination of \emph{distant supervision} (event-level only labeling) and \emph{weak supervision} (sparse and high-level labeling).

\subsection{Multimodal CED Dataset}\label{sec:ce_dataset}
\begin{table}[t]
\centering
\caption{Complex Event Classes and Definitions}
{\scriptsize
\begin{tabular}{@{}p{1.5cm}p{5.3cm}p{1cm}@{}}
\toprule
\textbf{Complex Events}   & \textbf{Definitions}   & \textbf{Category}                 \\ \midrule
Default ($e_0$)     & When no complex events of interest take place.                   \\ \midrule
Workspace sanitary protocol violation ($e_1$)   & A violation occurs if a person starts working (click or type) without 20 seconds of consecutive handwashing after using the restroom. Upon violation, trigger an alert and reset the system. & Sequential + Temporal             \\ \midrule
Sanitary eating habit violation ($e_2$)        & Hands are not cleaned if no 20-second consecutive handwashing occurs within 2 minutes before a meal session. A meal session starts when eating or drinking begins and ends when any activity other than eat, drink, or sit is detected.          & Sequential + Temporal             \\ \midrule
Inadequate brushing time ($e_3$)           & Brushing teeth for less than 2 minutes. If brushing stops, wait for 10 seconds; otherwise, report violation and reset the system.                                                                                                                        & Temporal (Relative + Duration)    \\ \midrule
Routine Sequence ($e_4$)           & brush $\rightarrow u^* \rightarrow$ eat $\rightarrow u^*\rightarrow$ drink $|$ brush $\rightarrow u^* \rightarrow$ drink $\rightarrow u^*\rightarrow$ eat, \newline where $u = A \setminus \{\text{brush, eat, drink}\}$\dag.               & Sequential - Relaxed              \\ \midrule
Start working and then take a break ($e_5$)        & sit $\rightarrow u^* \rightarrow$ type/click $\rightarrow v^* \rightarrow$ walk, \newline where $u = A\setminus\{\text{sit, type, click, walk}\}$, and $v = A\setminus\{\text{type, click, walk}\}$\dag.                & Sequential - Relaxed              \\ \midrule
Sufficient Washing Reminder ($e_6$)                 & When washing lasts for 30 seconds consecutively.                           & Temporal - Duration               \\ \midrule
Adequate brushing time ($e_7$)             & When brushing lasts a total of 2 minutes. The timer pauses if brushing stops but resumes if brushing restarts. Once the 2-minute threshold is reached, the event is reported, and the timer resets.                                      & Temporal (Relative + Duration)    \\ \midrule
Post-Meal Rest ($e_8$)                      & After eating, wait for at least 3 minutes to work.      & Temporal - Relative               \\ \midrule
Active Typing Session ($e_9$)          & The event occurs if at least 3 typing sessions (start typing, stop typing) happen within 60 seconds of the first session's start.                                                  & Repetition - Frequency            \\ \midrule
Focused Work Start ($e_{10}$)             & The event is triggered by sitting after being seated, as long as no walking occurs during this time. The event is reported after exactly 5 clicks after sitting and before walking.     & Repetition - Contextual           \\ \bottomrule
\end{tabular}
}
\label{tab:complex_events}

\parbox{0.95\linewidth}{%
\raggedright 
\footnotesize
\textbf{Notes:} \dag Here $A$ represents the set of all \emph{atomic events}.
}
\vspace{-1em}
\end{table}

As no large-scale dataset exists for online CED, we develop a multimodal dataset in a smart health monitoring setting, leveraging multimodal sensors for rich contextual information in realistic \emph{CE} tasks. The dataset includes 10 \emph{CE} classes spanning various categories, with detailed rule definitions in Table~\ref{tab:complex_events}.  

\subsubsection{Sensor Data.} To generate \emph{CE} sensor traces, we built a stochastic simulator that mimics daily human behaviors by generating a random \emph{AE} label every 5-second window following realistic distributions. We use 9 \emph{AE} classes: ``walk'', ``sit'', ``brush'', ``click mouse'', ``drink'', ``eat'', ``type'', ``flush toilet'', and ``wash''. To ensure the generated AE sequences are consistent with the \emph{CE} patterns, we define \emph{AE} transitions between windows. For example, to mimic the scenario where a person forgets to clean hands after using the restroom, ``wash'' has a 0.7 probability after ``flush\_toilet''. We also include traces without CE patterns to balance the dataset and prevent biased \emph{CE} rule learning, ensuring the model is exposed to scenarios with or without each \emph{CE}.  

\subsubsection{Labeling.} These \emph{AE} traces are converted to sensor traces by sampling the corresponding segments reflecting the activity from \textbf{WISDM}~\cite{wisdm} for IMU data and \textbf{ESC-70}~\cite{kitchen20} for audio data. Ground-truth \emph{CE} labels are generated using human-defined finite state machines (FSMs), one for each \emph{CE} class. The simulator first generates \emph{AE} labels for each sequence, which are then processed by FSMs to determine the corresponding online \emph{CE} labels.

\subsubsection{Training \& Test Data.}
The dataset consists of 5-minute \emph{CE} sensor traces synthesized using inertial and audio data from multiple subjects, with each \emph{CE} sequence containing 60 windows ($5min \times 60sec/min \div 5~sec = 60$). We generated 10,000 training and 2,000 validation examples, including sequences with no \emph{CE} occurrences to reflect real-world sparsity. The test dataset uses sensor traces from an unseen subject and includes 2,000 examples of 5-minute sequences. Additionally, we created out-of-distribution (OOD) datasets with 2,000 examples each for 15-minute and 30-minute sequences, maintaining the same \emph{CE} patterns but with longer \emph{AE} durations and wider temporal gaps  between key \emph{AEs} to introduce additional challenges for generalization.

\section{Online CED Pipeline Overview}
We propose a two-module online processing system as shown in Fig.~\ref{fig:ced-pipeline}. The system uses a non-overlapping sliding window of size $W$ (5 seconds) to segment raw sensor streams into data segments $\{s_t\}_{t=1}^T$, which are then processed by the following modules.

\begin{figure}[t]
    \centering
    \includegraphics[width=1\columnwidth]{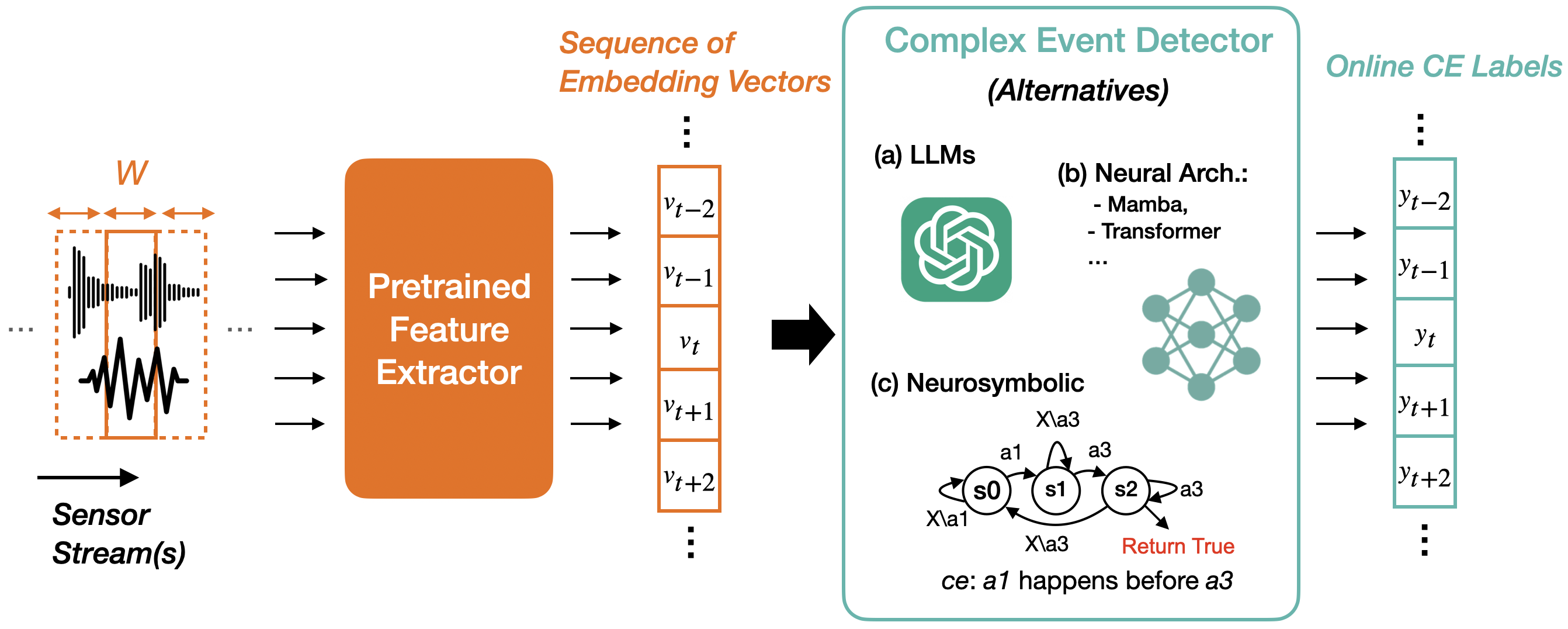}
    \caption{Overview of the online CED pipeline.}
    \Description{}
    \label{fig:ced-pipeline}
    \vspace{-1em}
\end{figure}

\subsection{Pretrained Feature Encoder}  
This module encodes each segment $s_t$ into a high-dimensional embedding vector $v_t$, producing a sequence of embedding vectors $\{v_t\}_{t=1}^T$. The encoder is pretrained to extract latent features from raw data and then frozen to ensure consistent embeddings across all downstream complex event detector models, enabling fair comparisons on \emph{CE} pattern reasoning. For our multimodal CED task, we develop a multimodal encoder for $s_t$ using early fusion. We use BEATs~\cite{chen2022beats} and LIMU-Bert~\cite{limubert} to extract audio and IMU features, which are then processed through separate GRU layers, concatenated and fused in a fusion layer to form a 128-dimensional embedding $v_t$. An MLP layer is added as the \textit{AE} classifier head for supervised training with cross-entropy loss. The encoder is then frozen, and the joint embedding $v_t$ is used for \emph{CE} detectors.

\subsection{Complex Event Detector}  
This module takes the embedding vectors \(\{v_t\}_{t=1}^T\) and detects \emph{CE} patterns, producing an online \emph{CE} label sequence \(\{y_t\}_{t=1}^T\), where \(y_t\) is the CE label at window \( t \). This component supports \emph{\textbf{LLM-based, neural, and neurosymbolic alternatives}} for comparison. Let the pretrained feature encoder be \(m\), and the complex event detector be \(g\). The online CED task objective defined in Eq.~\ref{eq:1} becomes:
\begin{equation}
    \min |\hat{y_t} - y_t|, \quad \textrm{where } \hat{y_t} = g\left(m\left(\mathbf{D}_t\right)\right), \quad 1 \leq t \leq T.
\end{equation}
Here, the feature encoder \(m\) remains fixed, while the complex event detector \(g\) is varied to compare different methods and architectures.

\section{Exploring LLMs as \emph{CE} Detectors}  
\subsection{Simplified Task} 
We evaluate LLMs as the complex event detector $g$ using ground truth \emph{AE} sequences as inputs, replacing the encoder $m$ with a perfect \emph{AE} classifier. This \emph{\textbf{simplifies}} the CED task by hiding sensor noise to focus solely on \emph{CE} reasoning. We further simplify by using only three \emph{CE} ($e_1$, $e_2$, and $e_3$). Each \emph{AE} label represents the activity in a 5-second window (e.g., $[walk", walk", ``sit"]$ means walking for 10 seconds and then sitting). These events occur over 5-minute intervals, yielding 60 \emph{AE} labels per example. LLMs also receive the CED task context and \emph{CE} definitions from Table~\ref{tab:complex_events}, and predict a \emph{CE} label every 5-second window. We evaluate 2,000 examples.\footnote{Prompts available: \href {https://anonymous.4open.science/r/LLM-CED-Prompts-CD6C/}{/r/LLM-CED-Prompts-CD6C/}}

\subsection{Evaluation} 
\subsubsection{Metrics.} We evaluate LLM performance using three metrics:
\begin{itemize}[leftmargin=10pt,nosep]
    \item \emph{Length Accuracy}: Measures the match rate between the length of the predicted \emph{CE} labels and input \emph{AE} sequences in the \emph{``n-to-n"} sequence prediction task required for the Online CED task.
    \item \emph{Conditional $F1$ Score}: Calculates element-wise $F1$ score for complex event labels, assessing both event type and timing accuracy, conditioned on sequences with the correct length $T$ and averaged over timestamps 1 to $T$.
    \item \emph{Coarse $F1$ Score}: Evaluates \emph{CE} labeling at a high level by checking if the correct event type is recognized within a 5-minute sample, without requiring a precise timestamp match.
\end{itemize}
We test LLMs on both zero-shot and few-shot tasks, with three input-output examples included in the prompt for few-shot experiments. 

\begin{table}[t]
\caption{Evaluation results of LLMs.}
    \begin{center}
    \small
    \setlength\tabcolsep{1.5pt}%
    \begin{tabular}{@{}lccccccccccc@{}}\toprule 
    & \multirowcell{2}{\makecell[c]{$Length$\\$Acc.$}}  && \multicolumn{4}{c}{$Coarse$ $F1$}  && \multicolumn{4}{c}{$Conditional$ $F1$}\\
    \cmidrule{4-7} \cmidrule{9-12}
    & && $e_1$ & $e_2$ & $e_3$ & Avg.&&  $e_1$ & $e_2$ & $e_3$ & Avg.\\ \midrule
    \textit{Zero-shot} &\\
    Qwen2.5-14B & 0.12 && 0.60 & 0.66 & 0.57 & 0.61 && 0.14 & 0.15 & 0.04 & 0.11\\
    GPT-4o-mini  & 0.04 && 0.0 & 0.0 & 0.64 & 0.21 && 0.0 & 0.0 & 0.0& 0.0 \\
    GPT-4o & 0.12 && 0.87 & 0.78 & 0.80 & 0.82  &&  0.14  & 0.59  & 0.03  & 0.25 \\
    o1-mini  & 0.13 && 0.78 & 0.84 & 0.80 & 0.80 && 0.0 & 0.72 & 0.02 & 0.25 \\
    o3-mini & 0.33 && 0.90 & 0.85 & 0.88 &  0.87 && 0.05  & 0.78 & 0.24  & 0.35 \\
    \midrule
    \textit{Few-shot} ($k = 3$) & \\
    Qwen2.5-14B & 0.14 && 0.62 & 0.66 & 0.60 & 0.63 && 0.13 & 0.13 & 0.03 & 0.10\\
    GPT-4o-mini & 0.03 && 0.0 & 0.0 & 0.58 & 0.19 && 0.0 & 0.0 & 0.0& 0.0 \\
    GPT-4o & 0.16 && 0.87 & 0.81 & 0.81 & 0.83 &&  0.13 & 0.63  & 0.14  & 0.30   \\
    o1-mini  & 0.25 &&  0.81 & 0.84 & 0.82 & 0.82 && 0.0 & 0.71 & 0.17 & 0.29 \\
    o3-mini & \textbf{0.44} && 0.94 &  0.86 & 0.88 & \textbf{0.89}  && 0.33  & 0.84 & 0.35  & \textbf{0.50} \\
    \bottomrule
    \end{tabular}
    \label{tab:results}
    \end{center}
    \vspace{-1em}
\end{table}

\subsubsection{Results.} We evaluated five SOTA LLM models on complex events 1, 2, and 3, as shown in Table~\ref{tab:results}. All models performed poorly, even on the simplified CED task. The low length accuracy suggests hallucination in long-chain reasoning, as LLMs often generated sequences between 55 and 65 labels instead of the required 60. Since the $F1$ score cannot be computed when sequence lengths mismatch, we calculated the conditional $F1$ score only for outputs with the correct length. Among zero-shot models, o3-mini performed slightly better but remained unsatisfactory. We also computed the coarse $F1$ score, which only evaluates whether a \emph{CE} occurred within a 5-minute window without requiring exact timing. In this relaxed setting, LLMs showed significant improvement, suggesting they can partially recognize \emph{CE} patterns but struggle with precise timing. Few-shot examples improved performance for stronger reasoning models like o3-mini but had little effect on others, indicating that o3-mini better utilizes \emph{CE} examples for self-checking its understanding. Although we did not evaluate o1 and o3 models, which may have stronger reasoning abilities, their high cost and inference time make them impractical for online CED, which requires frequent, low-latency API requests. In summary, while LLMs show potential for online CED, they suffer from hallucinations, poor long-chain reasoning, and high latency due to their transformer-based architecture, leading to delays in real-time inference for longer sensor traces.

\section{Neural and Neurosymbolic Methods}  
Given the limitations of LLMs, we design and train various neural and neurosymbolic architectures as alternatives for the complex event detector $g$ following some requirements.

\subsection{Architectures}
\subsubsection{Requirements}
\textbf{Causal Structure:} Online CED requires models to predict complex events at each time step $t$ using only past observations (0 to $t$) to ensure no future information is accessed.  \textbf{Minimum Receptive Field:} Neural networks must have a receptive field larger than the longest temporal patterns of complex events in the training data to capture full event patterns.  

\subsubsection{Variants}
We explore three types of architectures:  

\textbf{End-to-end Neural Architectures.} These models take high-dimensional sensor embeddings as input and are trained end-to-end for \emph{CE} detection. We compare: (1) a 5-layer \textbf{\textit{Unidirectional LSTM}} (hidden size 256, $\approx$ 2.5M parameters), (2) a \textbf{\textit{Causal TCN}}~\cite{bai2018tcn} with dilation rate 32, kernel size 3, and 128 filters per layer (8-min receptive field, $\approx$ 4.6M parameters),  (3) a 6-layer \textbf{\textit{Causal Transformer}} with triangular attention mask and positional encoding (hidden size 128, 8-head attention, $\approx$ 4.2M parameters), and (4) a 12-block \textbf{\emph{Mamba}} state-space model (hidden size 128, state size 64, $\approx$ 1.8M parameters). This configuration is chosen because the original Mamba paper~\cite{gu2024mamba} shows that two SSM blocks are equivalent to one Transformer layer.  

 \textbf{Two-stage Concept-based Architectures.} Denoted as \textbf{\emph{Neural AE + X}}, these models first use an \emph{\textbf{neural AE}} classifier to map each window of sensor embedding to an \emph{AE} class,  which are then processed by a neural backbone model \emph{\textbf{X}} for \emph{CE} detection. Variants include \emph{Neural AE + LSTM}, \emph{Neural AE + TCN}, \emph{Neural AE + Transformer}, and \emph{Neural AE + Mamba}. The backbone models are identical to those in the end-to-end architectures but take one-hot \emph{AE} class traces as input.

\textbf{Neurosymbolic Architecture:} We design a neurosymbolic model, \textbf{\emph{Neural AE + FSM}}, that integrates human knowledge of \emph{CE} rules. It uses the same neural \emph{AE} classifier and a user-defined symbolic reasoner, employing an FSM for each complex event rule, identical to those used in \emph{CE} labeling. We also explored a probabilistic FSM in ProbLog~\cite{problog}, leveraging softmax embedding for probabilistic reasoning over \emph{CE} sequences. However, it showed only marginal improvements and was excluded due to design complexity and reliance on expert knowledge.


\subsection{Training Loss}  
The online CED task faces severe class imbalance due to the temporal sparsity of \emph{CE} labels, where the majority class is ``0" (no event), dominating the label distribution. To address this, we use Focal Loss (FL)~\cite{DBLP:journals/corr/abs-1708-02002}, which emphasizes errors on rare but critical classes. The loss is defined as:  
\begin{equation}  
\min_\theta L_{FL}\left(\theta\right)=-\sum_{i=1}^{N}\sum_{t=1}^{T}\alpha_{y_i(t)}\left(1-p_{y_i(t)}\right)^\gamma \log \left(p_{y_i(t)}\right),
\end{equation}  
where $p_{y_i(t)}$ is the predicted probability of class $y_i$ at time $t$, $\gamma$ reduces the impact of frequent classes, and $\alpha_y$ balances class weights. After a hyperparameter grid search, we set $\gamma=2$, $\alpha_0=0.005$ for the most frequent class ``0'', and $\alpha_y=0.25$ for rare but critical \emph{CE} classes.

\subsection{Evaluation}  

\subsubsection{Experimental Setup.}  
All neural models are trained using the AdamW optimizer with Focal Loss, a learning rate of \(1 \times 10^{-3}\), weight decay of 0.1, and batch size of 256. Early stopping is applied based on validation loss, with a maximum of 5000 training epochs. Results are averaged over 10 random seeds. The \emph{Neural AE} classifier used for \emph{Neural AE + X} models is trained with the pretrained feature encoder, achieving 95\% \emph{AE} classification accuracy on the test set.

\subsubsection{Metrics.}  
We evaluate performance using the $F1$ score for each \emph{CE} class $e_i$ and report \emph{Positive $F1$} ($F1\_pos$), which is the average $F1$ over positive event classes ($e_1$ to $e_{10}$), excluding the less important ``negative'' label $e_0$. A higher $F1$ score indicates a better precision-recall balance, reflecting correctness and completeness.

\subsubsection{Results.}  
We evaluate model performance on \emph{CE}s defined in Table~\ref{tab:complex_events} across different training set sizes, as shown in Fig.~\ref{fig:ce_different_trainingsizes_boxplot}. The results indicate that \emph{\textbf{Mamba achieves the best performance}}, followed by LSTM. The \emph{Neural + X} models underperform compared to end-to-end models, likely due to errors and noise introduced by the \emph{Neural AE} classifier. This also explains why \emph{Neural AE + FSM} performs worse despite incorporating correct human-defined \emph{CE} rules. Additionally, we test model generalization on out-of-distribution (OOD) complex events lasting 15 and 30 minutes, which follow the same \emph{CE} rules but extend their temporal spans. As shown in Fig.~\ref{fig:ce_different_timespan_boxplot}, Mamba exhibits the best generalization among both end-to-end neural models and \emph{Neural AE + FSM}. Table~\ref{tab:baseline_different_temporal_span} further shows that \emph{\textbf{Mamba, when trained with more labeled sensor data on 5-minute sequences, improves generalization to longer unseen traces}}, suggesting that it efficiently learns \emph{CE} rules as training data increases. These findings indicate that Mamba may serve as a strong backbone for foundation models targeting long and complex temporal reasoning.

\begin{figure}[t]
    \centering
\includegraphics[width=0.95\columnwidth]{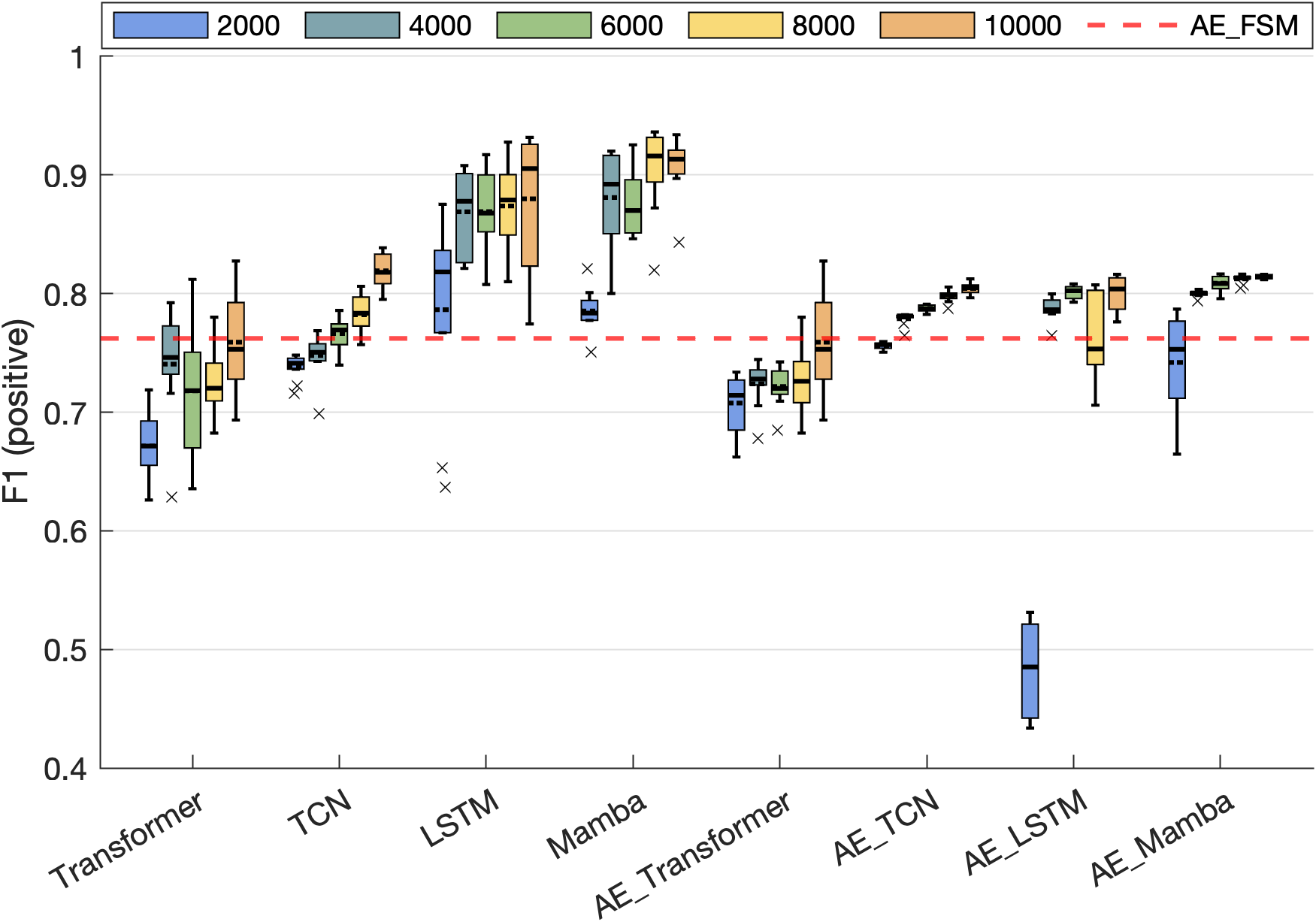}
    \caption{Boxplot of positive $F_1$ scores on complex events with different training data sizes. Solid line in the box shows median; dashed line in the box shows mean.}
    \Description{}
    \label{fig:ce_different_trainingsizes_boxplot}
    \vspace{-1em}
\end{figure}

\begin{figure}[t]
    \centering
\includegraphics[width=0.95\columnwidth]{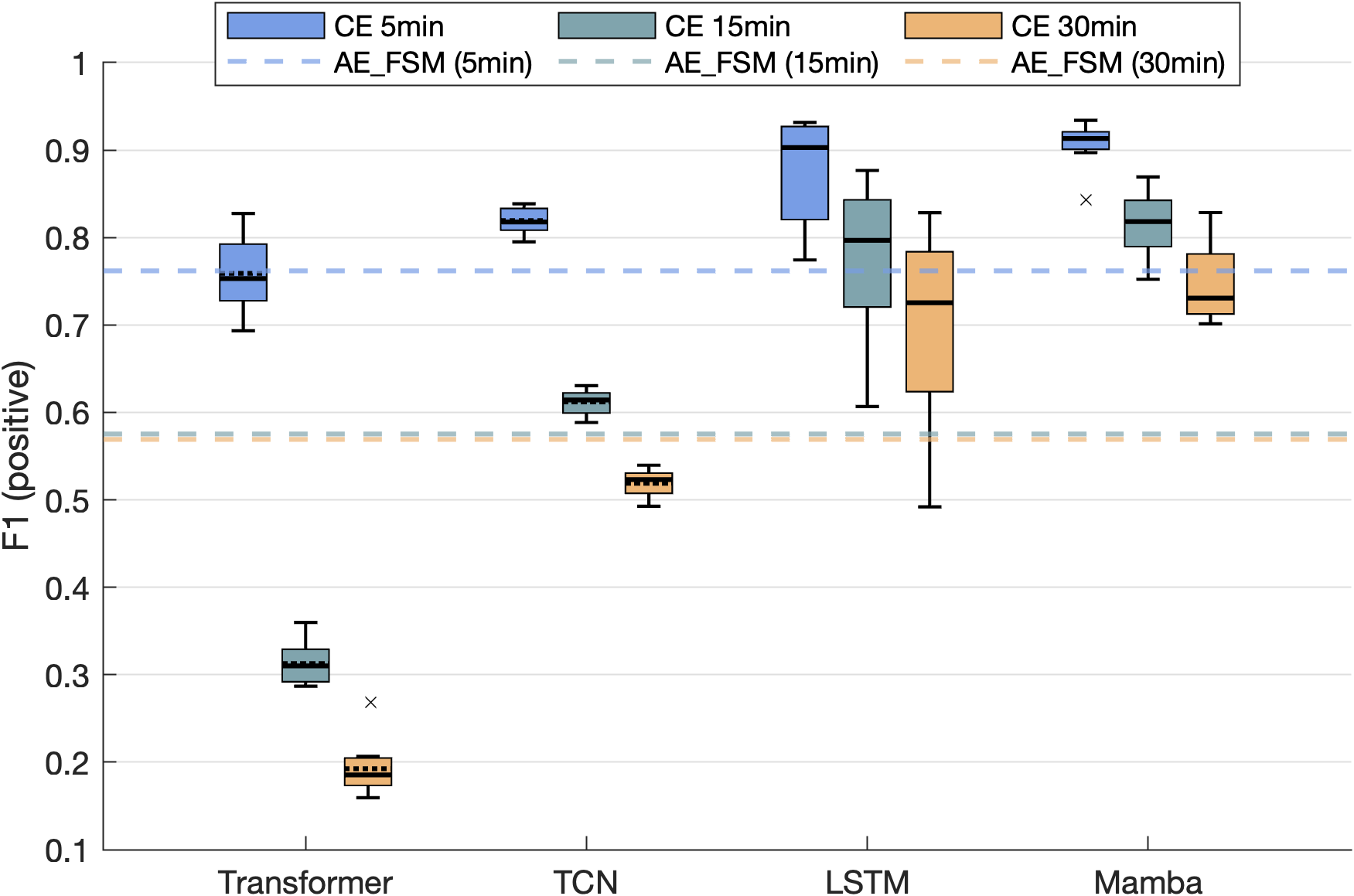}
    \caption{Positive $F1$ scores of different models tested on 5-min, 15-min, and 30-min \emph{CE} sensor data.}
    \Description{}
    \label{fig:ce_different_timespan_boxplot}
\end{figure}

\begin{table}[t]
    \centering
    \small
    \setlength{\tabcolsep}{4pt}
    \caption{Positive $F1$ scores of end-to-end Mamba model tested on 5-min, 15-min, and 30-min \emph{CE} sensor data (with a 2-sigma confidence interval). }
    \begin{tabular}{c c c c} 
        \toprule
         \textbf{Training} & \multicolumn{3}{c}{\textbf{Positive $F1$}}\\
        \cmidrule(lr){2-4}
         \textbf{Data Size} & \textbf{5min} & \textbf{15min (OOD)} & \textbf{30min (OOD)} \\
        \midrule
        2000 & .79 $\pm$ .04 & .68 $\pm$ .06 & .55 $\pm$ .07 \\
        4000 & .88 $\pm$ .08 & .78 $\pm$ .09 & .69 $\pm$ .14 \\
        6000 & .88 $\pm$ .06 & .78 $\pm$ .09 & .70 $\pm$ .09 \\
        8000 & .90 $\pm$ .08 & .81 $\pm$ .12 & .74 $\pm$ .13 \\
        10000 & .92 $\pm$ .06 & .82 $\pm$ .08 & .75 $\pm$ .09 \\
        \bottomrule
    \end{tabular}
    \label{tab:baseline_different_temporal_span}
\end{table}

\section{Conclusion}  
We formalized the online CED task and developed a dataset in a CPS-IoT scenario to evaluate various methods. Although LLMs showed potential, they failed on the simplified CED task due to poor precise reasoning over long temporal patterns. We then explored neural and neurosymbolic alternatives as \emph{CE} detectors. The end-to-end Mamba model outperformed all others, demonstrating smaller model size and better generalization to unseen longer \emph{CE} traces. This highlights its effectiveness in online CED tasks and its potential as a suitable backbone for complex temporal reasoning foundation models in CPS-IoT applications.

\begin{acks}
This research was sponsored in part by the DEVCOM ARL (award \# W911NF1720196), the AFOSR (awards \#  FA95502210193 and \# FA95502310559), the NIH (award \# 1P41EB028242), the NSF (award \# 2325956), and the EOARD (award \# FA8655-22-1-7017).
\end{acks}

\bibliographystyle{ACM-Reference-Format}
\bibliography{references}

\end{document}